\documentclass[runningheads]{llncs}
\usepackage{graphicx}

\usepackage{cite}
\usepackage{amsmath,amssymb,amsfonts}
\usepackage{algorithmic}
\usepackage{textcomp}
\usepackage{xcolor}
\usepackage{hyperref}

\usepackage{verbatim}
\usepackage{color, colortbl}
\definecolor{Green}{rgb}{0.6,1,0.6}
\usepackage{wrapfig, lipsum}
\usepackage{multirow}
 \usepackage{paralist}
\usepackage{microtype}

\begin{document}
\title{Learning Flexible Translation between Robot Actions and Language Descriptions}
\titlerunning{Learning Flexible Translation between Action and Language}
%
\author{Ozan Özdemir \and
Matthias Kerzel \and
Cornelius Weber \and \newline Jae Hee Lee \and Stefan Wermter
}
\authorrunning{O. Özdemir et al.}
%
\institute{Knowledge Technology, Department of Informatics, University of Hamburg \\
\email{\{ozan.oezdemir,matthias.kerzel,cornelius.weber,\newline jae.hee.lee,stefan.wermter\}@uni-hamburg.de} \newline
\url{www.knowledge-technology.info}}
\maketitle              
\begin{abstract}
Handling various robot action-language translation tasks flexibly is an essential requirement for natural interaction between a robot and a human. Previous approaches require change in the configuration of the model architecture per task during inference, which undermines the premise of multi-task learning. In this work, we propose the paired gated autoencoders (PGAE) for flexible translation between robot actions and language descriptions in a tabletop object manipulation scenario. We train our model in an end-to-end fashion by pairing each action with appropriate descriptions that contain a signal informing about the translation direction. During inference, our model can flexibly translate from action to language and vice versa according to the given language signal. Moreover, with the option to use a pretrained language model as the language encoder, our model has the potential to recognise unseen natural language input. Another capability of our model is that it can recognise and imitate actions of another agent by utilising robot demonstrations. The experiment results highlight the flexible bidirectional translation capabilities of our approach alongside with the ability to generalise to the actions of the opposite-sitting agent. 

\end{abstract}
\section{Introduction}
Learning language involves multiple modalities such as audio, vision and proprioception. For example, a colour word refers to a visual concept; sensing the weight of an object is related to the concept of force; the concept of position such as left and right can be learnt with proprioception and vision. More modalities can be enumerated that help with learning language but the essential component of language learning pertains to embodiment (i.e. having a body and interacting in the environment) \cite{bisk-etal-2020-experience}. The embodied language learning or language grounding has recently been a topic of interest at the crossroads between natural language processing (NLP) and robotics \cite{bisk-etal-2020-experience, abramson2020imitating, shridhar2018interactive, lynch2021language}. \begin{wrapfigure}{r}{4.5cm}
\includegraphics[width=4.5cm]{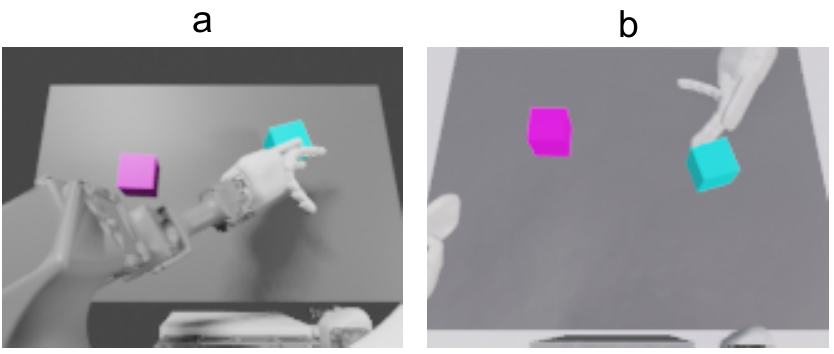}
\caption{Our object manipulation scenario: a) an example action (`push cyan slowly') undertaken by NICO \cite{kerzel2017nico} b) the same action undertaken by the opposite agent as seen by NICO.} \label{scenario}
\end{wrapfigure} Inspired by the early language development in children in which interactions in the environment are paired with language, language grounding approaches have achieved learning representations, forming of abstractions, sequence-to-sequence learning and bidirectional learning of action and language. However, these approaches are not designed to endow a robot with the autonomy to understand and choose the appropriate action to carry out during an interaction with a human. They are either designed to carry out a single task such as recognising an instruction and executing it \cite{hatori2018interactively, shridhar2018interactive, shao2020concept2robot, lynch2021language} or they can handle multiple tasks but they require the task mode in advance to know what is expected of them \cite{ogata2007parametricbias, yamada2018paired, Oezdemir_2021_ICDL}. In contrast, a truly autonomous agent must be able to decide whether to produce language or execute an action according to the verbal instruction given by its human partner. Therefore, end-to-end multimodal and multi-task models, which do not require adaptation to new tasks by the experimenter, are desired.

\begin{figure}[h]
\includegraphics[width=\textwidth]{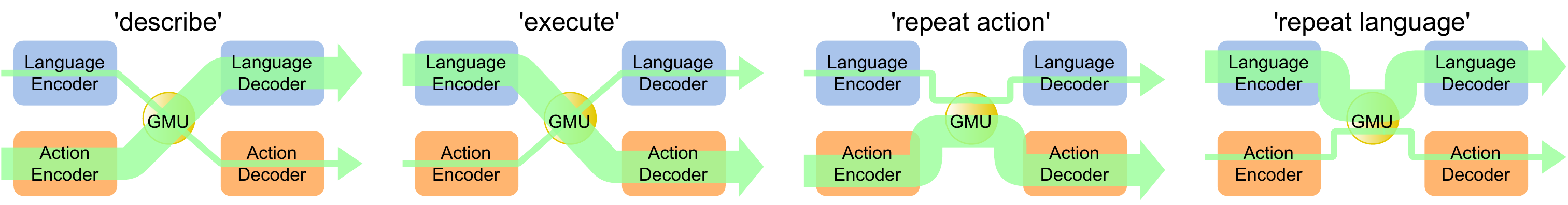}
\caption{The abstract architecture with four different tasks: `describe', `execute', `repeat action' and `repeat language'. The PGAE architecture consists of encoders, decoders and a GMU bottleneck. Each column shows our architecture with the information flow (indicated by the arrows) per task. The thick green arrows denote the main information flow in the respective signal's task.} \label{abstract_task}
\end{figure}

In this work, we address the problem of flexible bidirectional translation between robot actions and language. We define flexibility as translating between the two modalities without having to reconfigure the model for a specific task during inference. According to our scenario (Fig. \ref{scenario}), we expect our agent to flexibly translate from action to language and vice versa, viz. given textual descriptions, joint angle values and visual features, our agent must either manipulate an object or describe the object manipulation act carried by itself or the second agent depending on the situation. To this end, we introduce the paired gated autoencoders (PGAE) architecture for flexible translation between robot actions and language, realised by our humanoid robot NICO \cite{kerzel2017nico, kerzel2020teaching} in the simulation environment. PGAE includes an attention mechanism in its bottleneck which allows the model to directly exchange information between the action and language modalities. The attention mechanism, which is adopted from the gated multimodal units (GMU) \cite{arevalo2020gated}, acts as a filter to pick information between the modalities across all dimensions. Moreover, we signal the task by prepending a phrase to the language input and ensure that our model recognises the task and is trained accordingly. Thus, during inference, our model is able to do each of the translation tasks (Fig. \ref{abstract_task}) according to user input without having to configure the model in advance. Further, we test a realistic setup in which the NICO robot can describe and repeat the actions of the opposite-sitting second agent. Our experiment results show that PGAE performs competitively in terms of translations between language and action with the previous approaches \cite{yamada2018paired, Oezdemir_2021_ICDL} that implicitly bind the two modalities and in turn could use only some parts of the network, which is set a priori according to the given translation task. Our contributions can be summarised as:
\begin{compactenum}
    \item introducing an end-to-end neural network (NN) architecture that can flexibly handle various action-language translation tasks during inference, consistent with the training conditions,
    \item enabling the robot to recognise and imitate both the self-performed actions and the actions of an opposite-sitting agent.
\end{compactenum}

\section{Related Work}
Translation between language and action has been a topic of interest: there are approaches that learn the general mapping between objects and language as well as attributes like colour, texture and size \cite{hatori2018interactively, shridhar2018interactive}, and there exist approaches that learn complex manipulation concepts \cite{shao2020concept2robot, lynch2021language}. However, these approaches can only translate from language to action as their focus is not on bidirectional translation. Other approaches can translate from action to language \cite{heinrich2020, ELWW21}. Only few approaches \cite{ogata2007parametricbias, antunes2019multitimescale, yamada2018paired, abramson2020imitating, Oezdemir_2021_ICDL} are capable of bidirectional translation, i.e. they have the ability to translate a given action into language as well as to translate a given language description into an action. 

Abramson et al. \cite{abramson2020imitating} propose a complex paradigm combining supervised learning, reinforcement learning (RL) and imitation learning in order to solve the problem of intelligently interacting in an abstract 3D play environment while using language. In the environment, two agents communicate with each other as one agent (setter) asks questions to or instructs the other (solver) that answers questions and interacts with objects according to a given instruction. However, the scenario is abstract as the objects are interacted with unrealistically. Hence, proprioception is not used as the actions are abstract. Therefore, the transfer of the approach from simulation to the real world is non-trivial.

Ogata et al. \cite{ogata2007parametricbias} propose an RNN-based model to enable bidirectional translation between compound sentences and robotic arm motions. Artificial bias vectors are used to bind the two modalities, which have separate RNNs, to enable flexible translation between them. Yamada et al. \cite{yamada2018paired} introduce the paired recurrent autoencoders (PRAE) approach which can bidirectionally translate between robot actions and language in a one-to-one manner: each action has exactly one description. Similar to \cite{ogata2007parametricbias}, PRAE consists of independent action and language networks and uses a binding loss to align the hidden representations of the paired actions and descriptions. PRAE cannot handle one-to-many mapping between action and language, i.e. when an action can be described by multiple descriptions. Moreover, it cannot flexibly change the direction of the translation during inference. For instance, to translate from action to language, PRAE accepts joint values and visual features through its action encoder and uses its description decoder alone to output a description - it practically excludes its description encoder and action decoder. Inspired by the PRAE architecture, we proposed the paired variational autoencoders (PVAE) \cite{Oezdemir_2021_ICDL} that can translate between a robot action and its multiple descriptions - we enabled one-to-many translation between actions and descriptions by utilising the stochastic gradient variational Bayes-based sampling (SGVB) \cite{kingma2014auto} that randomises the hidden representation space so that descriptions that are equivalent in meaning are represented tightly together, whereas those that have different meanings are represented far from each other. Like PRAE, PVAE employs the binding loss to map descriptions and actions. Therefore, due to its artificial nature in its multimodality fusion, PVAE too must be in a certain configuration according to the desired translation direction. The aim of our work is therefore to lift this constraint and allow flexible use of the model triggered by a verbally provided signal.

\section{Proposed Method: PGAE}
\begin{figure}
\includegraphics[width=\textwidth]{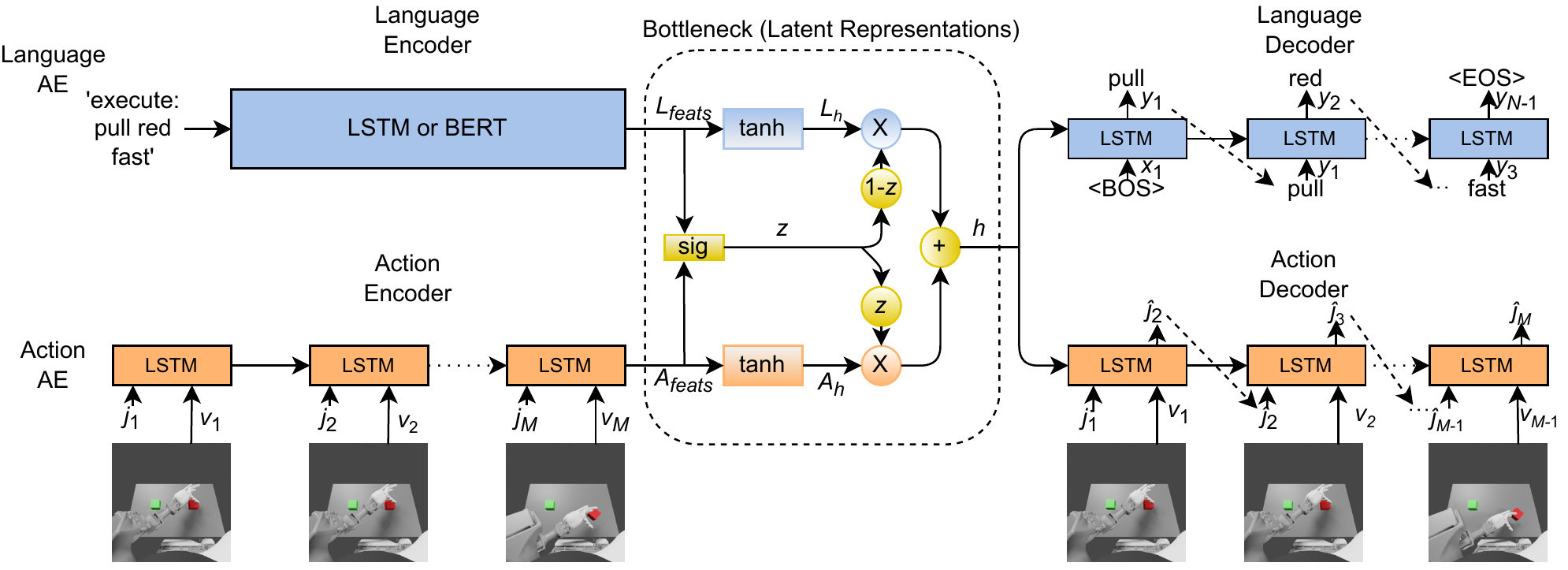}
\caption{The architecture of the PGAE model: Language encoder is either an LSTM or the BERT model. The action encoder and both decoders are LSTMs - we show unfolded versions of the LSTMs. The bottleneck, where the two streams are connected, is based on the GMU; $z$ is the gating vector, whilst $h$ is the shared representation vector.} \label{modelarch}
\end{figure}

Our paired gated autoencoders approach (PGAE) is a bidirectional translation model between robot actions and language. As can be seen in Fig. \ref{modelarch}, PGAE consists of two autoencoders, namely language and action. It is intended to associate simple robot actions like pushing a cube on the table with their corresponding language descriptions. PGAE accepts as input language descriptions, visual features extracted from images and joint angle values. PGAE outputs language descriptions and joint angle values conditioned on visual features. Moreover, PGAE is trained end-to-end with a signal prepended to the language input indicating the expected output of the training iteration. Five different signals are randomly chosen during training at each iteration: `describe', `execute', `repeat action', `repeat both' and `repeat language'.  

\subsubsection{Language Autoencoder}
The language autoencoder (AE) accepts as input one-hot encoded words of a description (or the whole description at once when BERT is used as language encoder) and produces a description by outputting a word at each time step. The language AE has an encoder, a decoder and hidden layers (in the bottleneck) that contribute to the common hidden representations. The language encoder embeds a description of length $N+1$, $(x_{1}, x_{2}, ..., x_{N+1})$, including the signal, into the final state $f_{N+1}$ as follows:
\begin{align*}
h_{t}^{\text{enc}}, c_{t}^{\text{enc}} &= \text{LSTM}(x_{t}, h_{t-1}^{\text{enc}}, c_{t-1}^{\text{enc}}) \hspace{5mm} (1\leq t\leq N+1),\\
f_{N+1}^{\text{enc}} &= [h_{N+1}^{\text{enc}}; c_{N+1}^{\text{enc}}],
\end{align*}
where 
$h_{t}^{\text{enc}}$ and $c_{t}^{\text{enc}}$ are the hidden and cell state of the LSTM at time step $t$ respectively. $h_{0}^{\text{enc}}$ and $c_{0}^{\text{enc}}$ are set as zero vectors, whereas $x_{1}$ is the signal word. The square brackets $[.;.]$ denote the concatenation operation. The LSTMs we use here and the action encoder and both decoders are a peephole LSTM \cite{sak2014long} following \cite{yamada2018paired, Oezdemir_2021_ICDL}. The language encoder LSTM can also be replaced by a pretrained language model to recognise unconstrained user instructions. Specifically, we use the pretrained BERT Base model \cite{devlin2019bert} as the language encoder. This variation of the model is called PGAE-BERT.

The language decoder autoregressively generates the descriptions word by word by expanding the shared latent representation vector $h$:
\begin{align*}
    h_{0}^{\text{dec}}, c_{0}^{\text{dec}} &= W^{\text{dec}} \cdot  h + b^{\text{dec}}, \\
    h_{t}^{\text{dec}}, c_{t}^{\text{dec}} &= \text{LSTM}(y_{t-1}, h_{t-1}^{\text{dec}}, c_{t-1}^{\text{dec}}) \hspace{3mm} (1\leq t\leq N-1), \\
    y_{t} &= \text{soft}( W^{\text{out}} \cdot h_{t}^{\text{dec}} + b^{\text{out}}) \hspace{5mm} (1\leq t\leq N-1),
\end{align*}
where $\text{soft}$ represents the softmax activation function. $y_{0}$ is the vector for the symbol indicating the beginning of the sentence, the $<$BOS$>$ tag. 
\subsubsection{Action Autoencoder}
The action autoencoder (AE) accepts a sequence of joint angle values and visual features as input and generates the appropriate joint angle values. It consists of an encoder, a decoder and latent layers (in the bottleneck) that contribute to the common latent representations. The action encoder encodes a sequence of length $M$, $((j_{1},v_{1}), (j_{2}, v_{2}), ..., (j_{M}, v_{M}))$ that is the combination of joint angles $j$ and visual features $v$. The visual features are extracted by the channel-separated convolutional autoencoder (CAE) in advance. The action encoder can be defined as\footnote{Some symbols in the equations coincide with the symbols used for the language AE.}:
\begin{align*}
h_{t}^{\text{enc}}, c_{t}^{\text{enc}} &= \text{LSTM}(v_{t}, j_{t}, h_{t-1}^{\text{enc}}, c_{t-1}^{\text{enc}}) \hspace{5mm} (1\leq t\leq M), \\
f_{M}^{\text{enc}} &= [h_{M}^{\text{enc}}; c_{M}^{\text{enc}}],
\end{align*}
where 
$h_{t}^{\text{enc}}$ and $c_{t}^{\text{enc}}$ are the hidden and cell state of the LSTM at time step $t$. $h_{0}^{\text{enc}}$, $c_{0}^{\text{enc}}$ are set as zero vectors. $f_{M}^{\text{enc}}$ is the final state of the action encoder.

The action decoder generates the joint angles at each time step by recursively expanding the shared latent representation vector $h$:
\begin{align*}
    h_{0}^{\text{dec}}, c_{0}^{\text{dec}} &= W^{\text{dec}} \cdot  h + b^{\text{dec}},\\
    h_{t}^{\text{dec}}, c_{t}^{\text{dec}} &= \text{LSTM}(v_{t}, \hat{\jmath}_{t}, h_{t-1}^{\text{dec}}, c_{t-1}^{\text{dec}})\hspace{5mm}(1\leq t\leq M-1), \\
    \hat{\jmath}_{t+1} &= \text{tanh}( W^{\text{out}} \cdot h_{t}^{\text{dec}} + b^{\text{out}}) \hspace{5mm} (1\leq t\leq M-1),
\end{align*}
where 
$\text{tanh}$ is the hyperbolic tangent activation function and $\hat{\jmath}_{1}$ is equal to ${j}_{1}$, i.e.\ joint angle values at the initial time step. Visual features $v$ are used as in teacher forcing.

\subsubsection{Bottleneck}
The language and action streams connect at the bottleneck, which is situated between the encoders and decoders of the model. We use a Gated Multimodal Unit (GMU) to fuse the language and action (joints, images) modalities. Thanks to its learned gating mechanism, GMU allows our model to flexibly learn multiple tasks according to the given command such as `describe' or `execute'. This way, our approach works in different translation directions using the whole model during inference without having to put the model in a specific mode. Our bottleneck can be defined with the following equations:
\begin{align*}
L_{\text{feats}} &= f_{N+1}, A_{\text{feats}} = f_{M}, \\
L_{h} &= \text{tanh}( W^{\text{L}} \cdot L_{\text{feats}} + b^{\text{L}}), \\
A_{h} &= \text{tanh}( W^{\text{A}} \cdot A_{\text{feats}} + b^{\text{A}}), \\
z &= \sigma (W^{z} \cdot [A_{\text{feats}}; L_{\text{feats}}] + b^{z}), \\
h &= z \odot A_{h} + (1-z) \odot L_{h},
\end{align*}
where $\sigma$ denotes the sigmoid activation function, whilst $\text{tanh}$ stands for the hyperbolic tangent activation function and $\odot$ is the Hadamard product. $h$ represents the shared hidden representation vector and is used as input to both language and action decoders.
\subsubsection{Signals}
Five different signals are used during training. Four of them can be used during inference. According to the given signal, the input and output of the model change.
\begin{itemize}
  \item \textbf{Describe} tells the model to describe the given action sequence, i.e. action-to-language translation. With this signal, the model accepts as input the sequence of visual features and joint angle values for the action as well as the $<$EOS$>$ tag for language. The model is then trained to output the correct description and the static joint angle values corresponding to the final time step of the action sequence.
  \item \textbf{Execute} signals the model to execute the given language description, i.e. language-to-action translation. With this signal, the model expects to be fed with the whole description sentence, and joint angle values and visual features corresponding to the first time step of the action sequence. The model is then expected to output the joint angle values of the action sequence from the action decoder and the $<$EOS$>$ from the language decoder.
  \item \textbf{Repeat Action} is the signal for reconstructing the sequence of joint angle values. PGAE expects as input the sequence of joint angle values and visual features for the action encoder and the $<$EOS$>$ tag in addition to the signal for the language encoder. The action decoder reconstructs the joint values and the language decoder outputs the $<$EOS$>$ tag.
  \item \textbf{Repeat Both} signal demands the paired language and action input to be present. With this signal, PGAE is trained similar to PVAE \cite{Oezdemir_2021_ICDL} end-to-end. The language encoder accepts the full description in addition to the phrase `repeat both', whilst the action encoder accepts the corresponding action sequence (joint values and visual features). The language decoder and action decoder outputs the full description and joint angle values correspondingly. This signal is intended to be used only during training, since it is implausible to expect the robot to repeat an action and its description at the same time.
  \item \textbf{Repeat Language} is used for reconstructing the full description. The full description and the first time step of the action sequence are fed as input to the encoders. As output, the language decoder reconstructs the description and the action decoder outputs the joint angle values of the first time step.
\end{itemize}
\subsubsection{Visual Feature Extraction}
Following the previous work \cite{Oezdemir_2021_ICDL}, we employ the channel-separated convolutional autoeconder architecture (CAE) to extract the visual features from images captured by the NICO robot. Instead of processing all three channels together, we train an instance of the CAE for each colour channel (red, green and blue) -- i.e. channel separation. The channel separation technique has been shown to distinguish between the colours of the objects more accurately \cite{Oezdemir_2021_ICDL}. Each channel of $120 \times 160$ RGB images fed into the channel-separated CAE at a time. The channel-separated CAE consists of a convolutional encoder, a fully-connected bottleneck and a deconvolutional decoder. After training for the RGB channels separately, the channel-specific visual features are extracted from the bottleneck and then concatenated. The resulting visual features $v$ are used as input to PGAE. For more details of the channel-separated CAE, please refer to the PVAE paper \cite{Oezdemir_2021_ICDL}.
\subsubsection{Loss Function}
The overall loss is calculated by adding up the reconstruction losses, i.e. language loss and action loss. The language loss, $L_{\text{lang}}$, is defined as the cross entropy loss between input and output words, whereas the action loss, $L_{\text{act}}$, is defined as the mean squared error (MSE) between original and predicted joint values:
\begin{equation*}
L_{\text{lang}} = \frac{1}{N-1} \sum_{t=1}^{N-1}\left ( -\sum_{i=0}^{V-1}  w^{\left [ i \right ]} x_{t+1}^{\left [ i \right ]}\log y_{t}^{\left [ i \right ]}\right),
\end{equation*}
\begin{equation*}
L_{\text{act}} = \frac{1}{M-1} \sum_{t=1}^{M-1}\left \| j_{t+1} - \hat{\jmath}_{t+1}  \right \|_{2}^{2},
\end{equation*}
where $V$ is the vocabulary size, $N$ is the number of words per description, M is the sequence length for an action trajectory and $w$ is the weight vector used to counter the imbalance in the frequency of words. The overall loss is the sum of the language and action loss:
\begin{equation*}
L_{\text{all}} = \alpha L_{\text{lang}} + \beta L_{\text{act}}
\end{equation*}
where $\alpha$ and $\beta$ are weighting factors for language and action terms in the loss function. In our experiments, we set $\alpha$ and $\beta$ to 1. 
\subsubsection{Training Details}
To train PGAE and PGAE-BERT, we first extract visual features using our channel-separated CAE. The visual features are used to condition the actions depending on the cube arrangement, i.e., the action execution according to a given description is dependent on the position of the target cube. Both PGAE and PGAE-BERT are trained end-to-end. PGAE and PGAE-BERT are trained for 6,000 epochs with the gradient descent algorithm and Adam optimiser \cite{kingma2015adam}. In our experiments, $h$ has 50 dimensions, $x$ has 28 dimensions, $j$ has 5 dimensions, $N$ is equal to 5 and $M$ is 50 for fast and 100 for slow actions. We take the learning rate as $10^{-5}$ with a batch size of 6 samples after determining them as optimal hyperparameters. After a few trials, we have decided to freeze the weights of BERT during training as fine-tuning it reduces the performance of our model. Since BERT has millions of parameters, fine-tuning it leads to overfitting.

\section{Experiment Results}
We train and test our model on the paired robot actions and descriptions dataset \cite{Oezdemir_2021_ICDL} that involves 864 samples of sequences of images, joint values and textual descriptions. The dataset consists of simple manipulation of two cubes of different colours on the table by the humanoid NICO robot. The NICO robot is a child-size humanoid robot with a camera in each of its two eyes. The dataset was created using the Blender software\footnote{\url{https://www.blender.org/}}. According to our scenario, using its left arm, NICO manipulates one of the two cubes on the table for each sample utilising the inverse kinematics solver provided on Blender. Each sample includes a sequence of first-person view images and joint angle values from NICO's left arm alongside a textual description of the action. In total, the dataset includes 12 distinct actions, 6 cube colours, 288 descriptions and 144 patterns (action-description-cube arrangement combinations). We slightly vary the 144 patterns six times randomly in terms of action execution in simulation. Out of 864 samples, we exclude 216 samples that involve every unique description and action type and use them as the test set. By carefully selecting the test samples, we ensure that the combinations of descriptions, action types and cube arrangements in the test set are not seen during training. For more details on the dataset, please consult the PVAE paper \cite{Oezdemir_2021_ICDL}. Additionally, we introduce a second agent that does the same actions and include the resulting images from the NICO's perspective. We use the visual features extracted from these images as additional input, and randomly select between them and the images that show NICO doing the actions. These cases are shown in Table \ref{tab:results}, with the `-opposite' suffix for describing or repeating the action of the second agent, and with the `-self' suffix for describing or repeating the action of NICO. Therefore, PGAE-self and PGAE-opposite are the same model trained on the same dataset -- the former is tested with self-agent (NICO's own) actions, while the latter with the second-agent actions. PGAE and PGAE-BERT are trained on the previous dataset as PVAE \cite{Oezdemir_2021_ICDL}. We test the models on action-to-language, language-to-action, action-to-action and language-to-language translations as shown in Table \ref{tab:results}.

\begin{table}[h]
\centering
\caption{Translation results for test set. Green background denotes the main output of the respective task (e.g., description for the action-to-language translation). As the opposite-sitting agent data is irrelevant for `repeat language' and `execute' tasks, PGAE-self and PGAE-opposite share those results.}\label{tab:results}
\begin{tabular}[t]{|c|c|c|c|c|}
\hline
& \textbf{Describe} & \textbf{Repeat Lang.} & \textbf{Execute} & \textbf{Repeat Act.}\\
  & \textbf{Act.\textrightarrow Lang.} & \textbf{Lang.\textrightarrow Lang.} & \textbf{Lang.\textrightarrow Act.} & \textbf{Act.\textrightarrow Act.} \\
  \hline
 \textbf{Approach} & Descr. Acc. & Descr. Acc.  & Descr. Acc.  & Descr. Acc.\\
 \hline
PGAE & \cellcolor{Green}93.05\% & \cellcolor{Green}96.30\% & 100\% & 100\%\\
PGAE-BERT & \cellcolor{Green}94.91\% & \cellcolor{Green}99.07\% & 100\% & 100\%\\
\hline
PGAE-self & \cellcolor{Green}80.56\% &\cellcolor{Green} && 100\%\\
PGAE-opposite & \cellcolor{Green}65.28\% & \multirow[c]{-2}{*}{\cellcolor{Green}93.98\%}& \multirow[c]{-2}{*}{100\%} & 100\%\\
 \hline
 \textbf{Approach} & J. Err. (nRMSE) & J. Err. (nRMSE) & J. Err. (nRMSE) & J. Err. (nRMSE)\\
  \hline
PGAE & 0.23\%  & 0.37\% & \cellcolor{Green}0.44\% & \cellcolor{Green}0.44\%\\
PGAE-BERT & 0.21\%  & 0.33\%  & \cellcolor{Green}0.44\% & \cellcolor{Green}0.42\%\\
\hline
PGAE-self & 0.58\% && \cellcolor{Green} & \cellcolor{Green}0.89\%\\
PGAE-opposite & 2.40\% & \multirow[c]{-2}{*}{0.73\%} & \multirow[c]{-2}{*}{\cellcolor{Green}0.79\%} & \cellcolor{Green}0.80\%\\
 \hline
\end{tabular}
\end{table}

\subsubsection{Action-to-Language Translation}
PGAE and its variants use the `describe' signal as input to the language encoder to translate from action to language. The first result column of Table \ref{tab:results} reports the accuracy of predicted test descriptions for different approaches. In order for a generated description to be accepted as correct, all of its words must match the ground truth description according to our predefined grammar, i.e. action-colour-speed-$<$EOS$>$. Moreover, the second half of the table shows the normalised root-mean-square error (nRMSE) between the generated and ground truth joint values. We calculate nRMSE values by dividing the square root of the MSE (which is used as the action loss $L_{\text{act}}$) by the observed range of joint values. Accordingly, PGAE is competitive with the earlier approach, PVAE (achieves 100\%, not given in the table), that has to be set in the specific configuration, which includes using only the action encoder and language decoder while bypassing the language encoder and action decoder by avoiding feeding any language input and not outputting the final joint values. Therefore, PVAE cannot output joint values when it is configured to be used for action-to-language translation. Both PGAE and PGAE-BERT, however, recognise the task from the signal in the language input and generate both the description and the joint values for all different tasks. Both PGAE and PGAE-BERT achieve near perfect joint value prediction. Moreover, PGAE-BERT performs slightly better than PGAE in terms of description accuracy. However, its main advantage over PGAE is that it has the potential to recognise unconstrained natural language due to the use of a pretrained language model. 

In the setting where we demand the model to describe the action done by the opposite agent, our model achieves 65\% accuracy (PGAE-opposite). It achieves 80\% accuracy when describing NICO's own actions (PGAE-self). Moreover, the joint value error also increases slightly to over 2\% for the case opposite-agent case, whereas it is comparable with the original approaches for PGAE-self. The decrease in the action-to-language accuracy is expected as introducing the second-agent actions makes the problem more challenging, e.g., pulling an object by the second agent might be interpreted as pushing the object by NICO itself. Moreover, this is an extra capability demonstrated by our approach.

\subsubsection{Language-to-Action Translation}
PGAE and its variants use the `execute' signal prepended to the description for the translation from language to action. The description accuracy (whether $<$EOS$>$ is outputted by the language decoder) and nRMSE between predicted and ground truth joint values for language to action translation are given in Column `Lang.\textrightarrow Act.'. All of the approaches are able to generate near perfect joint values (less than 1\% nRMSE). However, PVAE is not trained to generate descriptions ($<$EOS$>$ in this case) when it is configured to execute descriptions (N.A.). This highlights the superiority of using signals as part of the language input and having a common hidden representation vector over the artificial use of a loss term to align two separate streams. Training PGAE with the demonstrations from the opposite-sitting agent does not significantly affect the action-to-language performance (0.79\% for PGAE-self/PGAE-opposite).

\subsubsection{Language-to-Language \& Action-to-Action Translations} 
PGAE and PGAE-BERT are competitive with PVAE (achieves 100\%, not given in the table) in terms of the description accuracy for language-to-language translation and slightly better in terms of the joint value prediction for the action-to-action translation. PVAE does not have the capacity to output joint values for language-to-language and descriptions for action-to-action translations, whereas PGAE and PGAE-BERT almost perfectly output the initial time-step joint values for language-to-language translation and achieves perfect description accuracy in action-to-action translation. Training PGAE with the additional opposite-sitting agent demonstrations slightly increases the joint value error in action-to-action translation for both the actions executed by NICO (PGAE-self) and by the second agent (PGAE-opposite).

\section{Conclusion}
We have introduced an end-to-end NN approach that can flexibly perform translation between robot actions and language descriptions in multiple directions, some of which involving both first-person actions and opposite-sitting agent actions. By integrating the task signal in the language input, our approach can recognise the given task and output the suitable descriptions and joint values during inference. Our approach, PGAE, exhibits competitive performance in all four translation tasks while having a consistent configuration across learning and inference. With the additional demonstrations from a second agent, our model can not only recognise and imitate its own actions but also the actions of the second agent despite the challenging nature of the task. To our knowledge, this skill set has not been modelled by previous approaches. In summary, PGAE can perform various translation tasks robustly without any change in the use of the architecture between learning and test time, which the previous approaches lacked, through its attention-based explicit multimodal fusion mechanism and the insertion of the task signal to the language input. Furthermore, the realism in our simulation promises sim-to-real transfer, which we will tackle in the future. Another avenue is to embed our model into a continuous human-robot dialogue framework in a closed-loop. Finally, we can utilise RL for more dexterous object manipulation with diverse ways to execute an action.

\section*{Acknowledgements}
The authors gratefully acknowledge support from the German Research Foundation DFG under Project CML (TRR 169).

%
%
%
%
\bibliographystyle{plain}
\bibliography{references}





\end{document}